\documentclass[letterpaper]{article} 
\usepackage{aaai2027}  
\nocopyright
\usepackage[hyphens]{url}  
\usepackage{graphicx} 
\usepackage{natbib}  
\usepackage{caption} 
\usepackage{algorithm}
\usepackage{algorithmic}
\usepackage{multirow}
\usepackage{booktabs}
\usepackage{multirow}
\usepackage{amssymb}
\usepackage{amsmath}
\usepackage{xcolor}
\usepackage{adjustbox}
\usepackage{newfloat}
\usepackage{listings}
\DeclareCaptionStyle{ruled}{labelfont=normalfont,labelsep=colon,strut=off} 
\floatstyle{ruled}
\newfloat{listing}{tb}{lst}{}
\floatname{listing}{Listing}

\usepackage{booktabs}

\title{Spiking Local Interaction and Adaptive Complementary Fusion \\for Spiking Transformers}
\author{
Dongcheng Zhao\textsuperscript{\rm 1}\equalcontrib,
Sicheng Shen\textsuperscript{\rm 1}\equalcontrib,
Zhenyu Yang\textsuperscript{\rm 1},
Zhiyuan Li\textsuperscript{\rm 1},
Jinyan Yu\textsuperscript{\rm 1},
Yongjian Wang\textsuperscript{\rm 1},
Tiechui Yao\textsuperscript{\rm 2},
Wenli Zhang\textsuperscript{\rm 2},
Tielin Zhang\textsuperscript{\rm 1}\corresponding
}
\affiliations{
\textsuperscript{\rm 1}
Center for Excellence in Brain Science and Intelligence Technology,\\
State Key Laboratory of Brain Cognition and Brain-inspired Intelligence Technology,\\
Institute of Neuroscience, Chinese Academy of Sciences\\
\textsuperscript{\rm 2}
China Electric Power Research Institute Co., Ltd. (CEPRI) \\
zhaodc@ion.ac.cn; zhangtielin@ion.ac.cn
}

\begin{document}

\maketitle

\begin{abstract}

Spiking Transformers model token interactions primarily through spiking self-attention (SSA). However, binary query and key representations map continuous similarities to sparse and discrete relation responses, which may suppress weak relations and limit the propagation of local spatial context. To address this limitation, we introduce Spiking Local Interaction (SLI) and Adaptive Complementary Fusion (ACF). SLI establishes an attention-independent pathway for direct information exchange among neighboring spiking tokens using lightweight depthwise--pointwise transformations. ACF integrates SSA and SLI through layer-specific, channel-wise coefficients that adaptively balance their contributions at different network depths. The proposed design preserves the original attention formulation and can be incorporated into different Spiking Transformer architectures with modest parameter overhead. Experiments on ImageNet-1K, CIFAR-10, CIFAR-100, CIFAR10-DVS, and ADE20K show consistent improvements across image classification, event-based recognition, and semantic segmentation. In particular, QKFormer with SLI and ACF achieves $84.37\%$ Top-1 accuracy on ImageNet-1K and $37.5\%$ mIoU on ADE20K, where the segmentation model is trained without ImageNet pretraining. Ablation studies and qualitative analyses further indicate that SSA and SLI capture complementary interaction patterns and that learnable fusion consistently outperforms fixed weighting.

\end{abstract}
\section{Introduction}
Spiking neural networks (SNNs) encode information as sparse binary events and maintain temporal states through neuronal dynamics, offering a biologically inspired framework for potentially energy-efficient computation on neuromorphic and edge hardware~\citep{maass1997networks,roy2019towards}. Advances in surrogate-gradient learning, residual architectures, normalization, and neuronal modeling have enabled the direct training of increasingly deep SNNs on challenging visual benchmarks and facilitated their extension beyond convolutional architectures~\citep{wu2018spatio,zheng2021going,eshraghian2023training,fang2023spikingjelly,yao2022glif,fang2021deep}.

Spiking Transformers combine spike-driven computation with the flexible token interaction mechanisms of Transformers~\citep{vaswani2017attention,dosovitskiy2020image}. Their attention modules typically encode query, key, and value features as binary spikes and replace Softmax normalization with spike-compatible relation operations~\citep{zhou2022spikformer}. Subsequent studies have improved attention computation, residual information flow, and hierarchical architectures, extending Spiking Transformers from image recognition to dense prediction, language modeling, and multimodal tasks~\citep{zhou2023spikingformer,sdtv1,lei2025spike2former,Zhang_2022_CVPR,zhang2025spike,bal2024spikingbert,shen2025step,pan2026spikingbrain2,song2026spikevla}.

Despite these advances, binary spike representations impose inherent constraints on token-relation modeling. Sparse firing reduces query--key co-activation, while spike discretization maps continuous similarities to a limited set of co-firing responses. As a result, weak relations may be suppressed, and the remaining interactions are represented with limited numerical resolution. These constraints are particularly relevant to visual representations, where neighboring tokens often belong to the same spatial structure even when their instantaneous query and key spikes exhibit weak or zero co-activation. Relying exclusively on attention-based relations may therefore limit the propagation of local spatial context. Existing methods address information loss in Spiking Transformers through richer spike representations, temporal interaction, and redesigned attention computation~\citep{shen2024tim,shen2026teformer,lee2025spiking,wang2026bipolar,xiao2025rethinking,guo2025spiking}. However, most operate within the attention pathway and remain dependent on the relations derived from query and key features. In contrast, the inherent spatial neighborhoods of visual tokens provide a complementary interaction structure that can be modeled independently of query--key co-activation.

To exploit this structure, we introduce Spiking Local Interaction (SLI) and Adaptive Complementary Fusion (ACF). SLI directly exchanges information among neighboring spiking tokens through lightweight depthwise--pointwise transformations, providing an attention-independent pathway for local spatial interaction. Spiking self-attention (SSA) and SLI capture token relations through complementary mechanisms. SSA models input-dependent relations, whereas SLI performs topology-constrained information exchange within spatial neighborhoods. ACF integrates the two pathways using layer-specific, channel-wise coefficients, allowing their relative contributions to adapt across network depth. The fused representations are propagated to subsequent blocks, enabling deeper attention layers to operate on features that incorporate both attention-based relations and local spatial context. The proposed design preserves the original attention formulation and can be incorporated into different Spiking Transformer architectures with modest parameter overhead.

Our main contributions are summarized as follows:

\begin{itemize}
\item We introduce Spiking Local Interaction (SLI), an attention-independent pathway that directly exchanges information among spatially neighboring spiking tokens, complementing the sparse and discrete relations modeled by spiking self-attention.

\item We propose Adaptive Complementary Fusion (ACF), which learns layer-specific, channel-wise coefficients to coordinate attention-based and local interactions while preserving the original attention formulation.

\item Extensive experiments on image classification, event-based recognition, and semantic segmentation show consistent improvements across different Spiking Transformer architectures with modest parameter overhead.

\end{itemize}

\section{Related Work}

Research on Spiking Transformers has primarily focused on scalable spike-driven architectures and enhanced token interaction mechanisms~\cite{zhao2026position}.

\subsection{Spike-Driven Transformer Architectures}

Spikformer established a directly trained Spiking Transformer framework by introducing Spiking Self-Attention, where query, key, and value representations are encoded as binary spikes and softmax normalization is removed~\citep{zhou2022spikformer}. Spikingformer subsequently redesigned the residual architecture to maintain spike-driven information flow in deeper networks~\citep{zhou2023spikingformer}. Spike-driven Transformer further reformulated attention as sparse masking and accumulation-based computation, reducing reliance on multiplication-intensive operations~\citep{sdtv1}.

Subsequent work has focused on improving scalability and architectural efficiency. Meta-SpikeFormer combines convolutional processing in shallow stages with spiking self-attention in deeper stages, enabling hierarchical feature extraction~\citep{sdtv2}. E-SpikeFormer introduces spike-firing approximation, sparse spiking convolution, and masked image modeling to facilitate large-scale training and asynchronous spike-driven inference~\citep{sdtv3}. QKFormer develops linear-complexity QK attention together with hierarchical spiking representations and efficient downsampling~\citep{zhou2024qkformer}. Hadamard spike interactions further simplify relation computation and reduce communication overhead~\citep{jiangefficient}. These advances have improved the scalability, efficiency, and spike-driven design of Spiking Transformers.

\subsection{Interaction Enhancement in Spiking Transformers}

A growing body of work has sought to improve the information represented and exchanged by spiking attention. Bipolar Spiking Attention employs ternary spike representations and Shiftmax to retain polarity-aware interactions~\citep{wang2026bipolar}. $\alpha$-SSA modifies similarity computation between sparse queries and keys by assigning adjustable contributions to matched non-spiking pairs~\citep{xiao2025rethinking}. $A^{2}OS^{2}A$ instead combines binary queries, real-valued keys, and ternary values to relax the representational constraints imposed by complete binarization~\citep{guo2025spiking}. Beyond attention representation, SEMM introduces spike-based routing across attention heads and channel experts~\citep{zhou2024spiking}, while SpiLiFormer applies biologically inspired lateral inhibition to suppress background responses and emphasize target-related activations~\citep{zheng2025spiliformer}.

Other studies enhance interaction along temporal, spatial, or frequency dimensions. TEFormer performs forward temporal fusion within attention and gated backward temporal modeling within the MLP~\citep{shen2026teformer}. STAtten jointly captures spatial and temporal dependencies within local temporal blocks~\citep{lee2025spiking}, whereas MD-Mixer incorporates multi-scale historical information through learnable temporal delays~\citep{shi2026temporal}. Max-Former restores local high-frequency information attenuated by the low-pass characteristics of spiking neurons~\citep{fang2026spiking}, and SWformer combines spiking wavelet transformations with local convolution to facilitate frequency-aware feature interaction~\citep{fang2024spiking}.

Most existing approaches enhance interaction by modifying attention computation, spike representations, or temporal and frequency modeling. In contrast, direct spatial interaction among neighboring spiking tokens through an explicit attention-independent pathway remains comparatively underexplored.

\begin{figure*}[!t]
    \centering
    \includegraphics[width=0.8\linewidth]{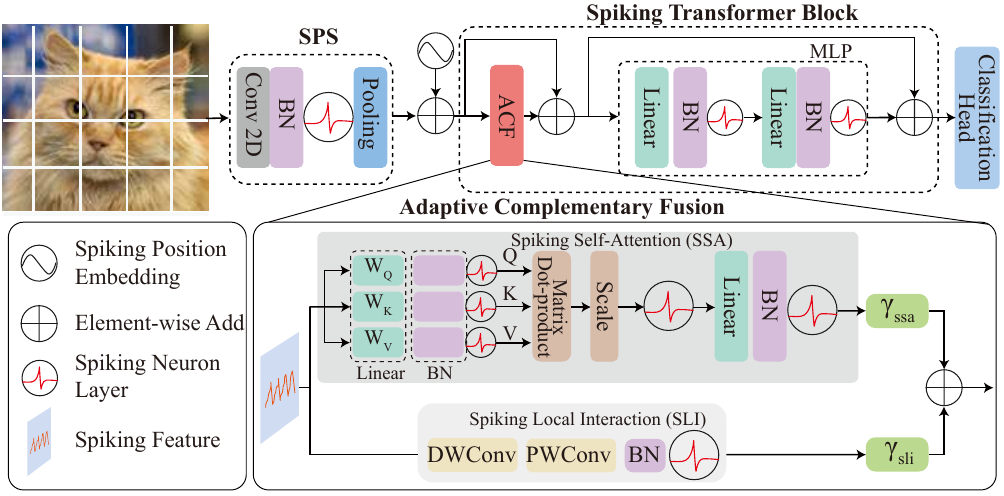}
\caption{
\textbf{Architecture of the proposed spiking Transformer with Adaptive
Complementary Fusion (ACF).}
Each block combines spiking self-attention (SSA) with a parallel Spiking
Local Interaction (SLI) pathway. SSA constructs input-dependent token
relations, while SLI aggregates local spatial context independently of
QK co-activation. Their responses are adaptively calibrated by
learnable channel-wise coefficients, and the fused representation is
propagated to subsequent blocks.
}
\label{fig:spiking_transformer_pipeline}
\end{figure*}

\section{Methods}

\subsection{Preliminaries}

\paragraph{Spiking Neuron.}
We employ the Leaky Integrate-and-Fire (LIF) neuron to model temporal
dynamics. At each time step $t$, the membrane integration, spike generation, and reset dynamics are formulated as
\begin{equation}
\left\{
\begin{aligned}
H^t &= \lambda V^{t-1} + I^t, \\
S^t &= \Theta\left(H^t - V_{\mathrm{th}}\right), \\
V^t &= H^t\left(1-S^t\right) + V_{\mathrm{reset}}S^t,
\end{aligned}
\right.
\label{eq:lif_dynamics}
\end{equation}
where $I^t$ denotes the input current, while $H^t$ and $V^t$ represent the membrane potentials before spike generation and after reset, respectively. The decay factor $\lambda\in[0,1]$ controls the retention of the membrane state from the preceding time step. Here, $V_{\mathrm{th}}$ and $V_{\mathrm{reset}}$ denote the firing threshold and reset potential. The Heaviside function $\Theta(\cdot)$ generates a binary spike
$S^t\in\{0,1\}$. During training, its non-differentiable derivative is
approximated using a surrogate gradient.

\paragraph{Spiking Patch Splitting.}
Spiking Patch Splitting (SPS) progressively transforms an input sequence
$\mathcal{I}\in\mathbb{R}^{T\times C\times H\times W}$ into spatiotemporal
tokens. For static images, the input is replicated across time steps, whereas
event streams are discretized into temporal frames. The $r$-th SPS stage is
defined as
\begin{equation}
Z_r^t
=
\operatorname{SN}
\left(
\operatorname{BN}
\left(
\operatorname{Conv}_r
\left(
Z_{r-1}^t
\right)
\right)
\right),
\qquad
Z_0^t=\mathcal{I}^t,
\label{eq:sps_stage_transformation}
\end{equation}
where $\operatorname{Conv}_r(\cdot)$, $\operatorname{BN}(\cdot)$, and
$\operatorname{SN}(\cdot)$ denote convolution, batch normalization, and
spiking-neuron activation, respectively. Successive SPS stages progressively
reduce the spatial resolution and increase the feature dimension.

After $R$ stages, the feature maps
$Z_R\in\mathbb{R}^{T\times D\times H'\times W'}$ are flattened along the
spatial dimensions and arranged as a token sequence:
\begin{equation}
X
=
\operatorname{Flatten}(Z_R)
\in
\mathbb{R}^{T\times N\times D},
\qquad
N=H'W',
\label{eq:sps_token_representation}
\end{equation}
where $T$, $N$, and $D$ denote the number of time steps, spatial tokens, and
embedding dimensions, respectively.

\subsection{Sparse and Discrete Relation Responses in Spiking Attention}

We first examine how spike-based attention differs from conventional
self-attention in constructing token relations. For clarity, we omit the
temporal dimension and multi-head decomposition and consider a single
attention head. Given a token sequence
$X\in\mathbb{R}^{N\times D}$, conventional self-attention is formulated as
\begin{equation}
\operatorname{Attn}(X)
=
\operatorname{Softmax}
\left(
\frac{QK^{\top}}{\sqrt{d}}
\right)V,
\label{eq:conventional_self_attention}
\end{equation}
where $Q=XW_q$, $K=XW_k$, and $V=XW_v$, and $d$ denotes the feature
dimension of each attention head. Continuous query--key correlations produce
fine-grained similarity scores, which are converted by Softmax into
normalized relative weights before value aggregation.

In spiking Transformers, the query, key, and value representations are
discretized by spiking neurons. Omitting normalization layers for brevity,
the projections in the $l$-th block are written as
\begin{equation}
\begin{aligned}
Q_l &= \operatorname{SN}\left(X_lW_{q,l}\right),\\
K_l &= \operatorname{SN}\left(X_lW_{k,l}\right),\\
V_l &= \operatorname{SN}\left(X_lW_{v,l}\right).
\end{aligned}
\label{eq:spiking_qkv_projections}
\end{equation}
Their relation responses can be generally expressed as
\begin{equation}
A_l
=
\Phi_l(Q_l,K_l),
\label{eq:general_spiking_attention_map}
\end{equation}
where $\Phi_l(\cdot,\cdot)$ denotes the relation construction rule of the
underlying backbone. Standard spiking self-attention (SSA) adopts
$\Phi_l(Q_l,K_l)=Q_lK_l^{\top}$, whereas QKFormer employs a different
spike-driven query--key interaction. Despite their different formulations,
both construct relation responses from sparse binary representations.

We use standard SSA to illustrate the influence of spike discretization on
relation construction. The response between the $i$-th query token and the
$j$-th key token is
\begin{equation}
a_{ij}^{l}
=
\sum_{r=1}^{d}
q_{i,r}^{l}k_{j,r}^{l},
\qquad
q_{i,r}^{l},k_{j,r}^{l}\in\{0,1\}.
\label{eq:spiking_cofiring_score}
\end{equation}
Thus, $a_{ij}^{l}\in\{0,1,\ldots,d\}$ measures the number of
query--key co-firing events. Under the simplifying assumptions of homogeneous
firing probabilities and independent query--key firing events, its expected
value and zero-response probability are approximated by
\begin{equation}
\mathbb{E}\left[a_{ij}^{l}\right]
\approx
d\,p_{q,l}p_{k,l},
\qquad
\Pr\left(a_{ij}^{l}=0\right)
\approx
\left(1-p_{q,l}p_{k,l}\right)^d,
\label{eq:ssa_sparse_response_analysis}
\end{equation}
where $p_{q,l}$ and $p_{k,l}$ denote the average firing probabilities of
$Q_l$ and $K_l$, respectively. 

\begin{figure}[!t]
    \centering
    \includegraphics[width=1.0\linewidth]{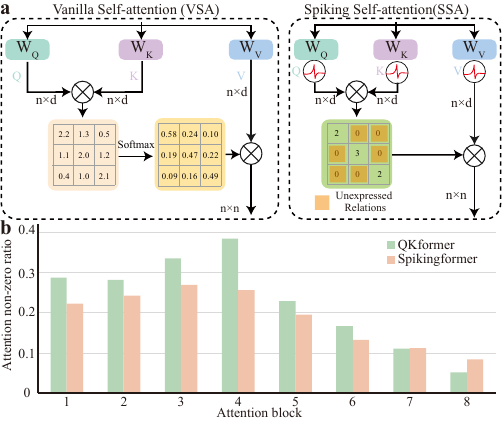}
\caption{
\textbf{Sparse and discrete relation responses in spiking attention.}
\textbf{(a)} Comparison of vanilla self-attention and spiking
self-attention.
\textbf{(b)} Attention-response non-zero ratios across eight blocks of
Spikingformer and QKFormer.
}
    \label{fig:ssa_information_attenuation}
\end{figure}

Sparse firing reduces query--key co-activation and produces more zero-valued
responses, while spike discretization compresses continuous similarities into
a finite set of integer-valued co-firing counts. As shown in
Fig.~\ref{fig:ssa_information_attenuation}, the proportion of non-zero
attention responses remains below $40\%$ across all eight blocks of both
Spikingformer-8-768 and QKFormer-8-768, and falls below $10\%$ in the final
block. These results indicate that sparse and low-resolution relation
responses persist across different spiking attention formulations, leaving
many potential token interactions unexpressed and potentially limiting the
propagation of local spatial structure. This motivates an interaction pathway
that directly propagates information among neighboring tokens without
depending on query--key co-activation.

\subsection{Spiking Local Interaction}

To provide an interaction pathway independent of query--key co-activation,
we introduce Spiking Local Interaction (SLI), which directly propagates
information among neighboring spiking tokens. Unlike SSA, whose
input-dependent relations are determined by query--key responses, SLI
operates over the predefined spatial neighborhoods of visual tokens.

Given $X_l\in\mathbb{R}^{T\times N\times D}$ with $N=H_lW_l$, SLI is
implemented using a lightweight depthwise--pointwise transformation:
\begin{equation}
\mathcal{S}_l(X_l)
=
\operatorname{SN}
\left(
\operatorname{BN}
\left(
\operatorname{PWConv}_l
\left(
\operatorname{DWConv}_l(X_l)
\right)
\right)
\right),
\label{eq:spiking_local_interaction}
\end{equation}
where the convolutions operate on the two-dimensional spatial layout of the
tokens at each time step. The depthwise convolution aggregates neighboring
responses independently within each channel, while the pointwise convolution
performs cross-channel projection. The aggregated responses are subsequently
normalized and re-encoded into binary spikes. By integrating neighborhood
responses before spike generation, SLI allows local evidence to accumulate
prior to thresholding, providing local structural context without relying on
query--key co-activation.

In a conventional spiking Transformer block, the attention branch updates
the token representation as
$\widehat{X}_l=X_l+\mathcal{A}_l(X_l)$, where
$\mathcal{A}_l(\cdot)$ denotes the SSA transformation. SSA and SLI capture
token interactions through distinct mechanisms. SSA models input-dependent
relations, whereas SLI aggregates topology-constrained local context. Their
relative contributions may vary across network depth and feature channels,
motivating an adaptive mechanism to coordinate the two pathways.

\subsection{Adaptive Complementary Fusion}

To coordinate the complementary interactions of SSA and SLI, we introduce
Adaptive Complementary Fusion (ACF). A fixed fusion rule cannot accommodate
variations in the relative importance of attention-based and local
interactions across network depth and feature channels. ACF therefore assigns
independent channel-wise scaling vectors to the two pathways.
\begin{equation}
\widetilde{X}_l
=
X_l
+
\gamma_{\mathrm{ssa},l}
\odot
\mathcal{A}_l(X_l)
+
\gamma_{\mathrm{sli},l}
\odot
\mathcal{S}_l(X_l),
\label{eq:adaptive_complementary_fusion}
\end{equation}
where $\mathcal{A}_l(\cdot)$ and $\mathcal{S}_l(\cdot)$ denote the SSA and
SLI transformations, respectively. The learnable scaling vectors
$\gamma_{\mathrm{ssa},l},\gamma_{\mathrm{sli},l}\in\mathbb{R}^{D}$
calibrate their contributions along the channel dimension and are broadcast
over the temporal and token dimensions. Both vectors are initialized to
$0.5$ and optimized jointly with the network, allowing each block to learn
channel-specific contributions from SSA and SLI.

The fused representation is subsequently processed by the standard residual
spiking MLP:
\begin{equation}
X_{l+1}
=
\widetilde{X}_l
+
\mathcal{M}_l(\widetilde{X}_l),
\label{eq:acf_mlp_update}
\end{equation}
where $\mathcal{M}_l(\cdot)$ denotes the spiking MLP transformation in the
$l$-th block. Since $X_{l+1}$ serves as the input to the subsequent block,
the corresponding query, key, and value spikes are generated from features
that have jointly incorporated attention-based relations and local spatial
context. This enriched representation enables deeper blocks to construct more informative query--key relations, supporting the progressive refinement of token interactions across network depth.

\section{Results}
\begin{table*}[!t]
\centering
\caption{
ImageNet-1K classification results.
$^*$ denotes our reproduction under the original experimental settings, and
$^\dagger$ denotes an input resolution of $288\times288$; all other models
use $224\times224$. HST denotes Hierarchical Spiking Transformer.
}
\label{tab:imagenet}

\begin{tabular}{llcccc}
\toprule
Methods & Type & Architecture & Params (M) & Time Steps & Top-1 Acc. (\%) \\
\midrule

ViT~\cite{dosovitskiy2020image}
& ANN & ViT-B/16 & 86.59 & 1 & 77.90 \\

\midrule

\multirow{2}{*}{DeiT~\cite{touvron2021training}}
& ANN & DeiT-B & 86.59 & 1 & 81.80 \\
& ANN & DeiT-B$^\dagger$ & 86.59 & 1 & 83.10 \\

\midrule



\multirow{2}{*}{Spikformer~\cite{zhou2022spikformer}}
& SNN & Spikformer-8-512 & 29.68 & 4 & 73.38 \\
& SNN & Spikformer-8-768 & 66.34 & 4 & 74.81 \\

\midrule

\multirow{2}{*}{SDT~\cite{sdtv1}}
& SNN & SDT-8-512 & 29.68 & 4 & 74.57 \\
& SNN & SDT-8-768$^\dagger$ & 66.34 & 4 & 77.07 \\

\midrule

\multirow{2}{*}{$\alpha$-SSA~\cite{xiao2025rethinking}}
& SNN & $\alpha$-SSA-ViT-8-512 & 29.70 & 4 & 75.39 \\
& SNN & $\alpha$-SSA-Swin-10-512 & 54.60 & 4 & 80.62 \\

\midrule

\multirow{2}{*}{$A^2OS^2$~\cite{guo2025spiking}}
& SNN & Spiking Transformer-8-512 & 29.68 & 4 & 76.28 \\
& SNN & Spiking Transformer-10-512 & 36.01 & 4 & 78.66 \\

\midrule

\multirow{2}{*}{SEMM~\cite{zhou2024spiking}}
& SNN & Spikingformer-8-384 & 16.05 & 4 & 73.58 \\
& SNN & Spikingformer-8-512 & 28.22 & 4 & 76.03 \\

\midrule

\multirow{2}{*}{STAtten+~\cite{lee2025spiking}}
& SNN & SDT-8-768 & 66.34 & 4 & 78.11 \\
& SNN & SDTV2-8-512 & 55.40 & 4 & 79.85 \\

\midrule

\multirow{2}{*}{BSA~\cite{wang2026bipolar}}
& SNN & HST-10-384 & 16.47 & 4 & 79.21 \\
& SNN & HST-10-512 & 29.08 & 4 & 82.39 \\

\midrule

\multirow{3}{*}{Spikingformer~\cite{zhou2023spikingformer}}
& SNN & Spikingformer-8-512 & 29.68 & 4 & 74.79 \\
& SNN & Spikingformer-8-768 & 66.34 & 4 & 75.85 \\

\midrule

\multirow{3}{*}{QKFormer~\cite{zhou2024qkformer}}
& SNN & HST-8-512* & 22.76 & 4 & 80.02 \\
& SNN & HST-10-512 & 29.08 & 4 & 82.04 \\
& SNN & HST-8-768* & 50.77 & 4 & 83.33 \\

\midrule

\multirow{2}{*}{Spikingformer+\textbf{SLI+ACF}}
& SNN & Spikingformer-8-512 & 31.84 & 4 & \textbf{75.96}\\
& SNN & Spikingformer-8-768 & 71.14 & 4 & \textbf{76.95}\\

\midrule

\multirow{3}{*}{QKFormer+\textbf{SLI+ACF}}
& SNN & HST-8-512 & 24.26 & 4 & \textbf{82.61}\\
& SNN & HST-10-512 & 31.12 & 4 & \textbf{83.33}\\
& SNN & HST-8-768 & 54.11 & 4 & \textbf{84.37}\\

\bottomrule
\end{tabular}

\end{table*}
\subsection{Results on ImageNet-1K Classification}

We evaluate the proposed SLI--ACF design on ImageNet-1K~\cite{deng2009imagenet} to assess its effectiveness in large-scale visual recognition. As shown in Table~\ref{tab:imagenet}, incorporating SLI and ACF consistently improves Spikingformer and QKFormer across all evaluated model scales. For QKFormer, the Top-1 accuracies of HST-8-512, HST-10-512, and HST-8-768 increase from 80.02\%, 82.04\%, and 83.33\% to 82.61\%, 83.33\%, and 84.37\%, respectively. Among the evaluated models, HST-8-768 achieves the highest accuracy with 54.11M parameters
at the default $224\times224$ resolution.
This result also exceeds the 83.10\% of
DeiT-B$^\dagger$, which uses 86.59M parameters and a $288\times288$ input.
Consistent improvements are also observed on Spikingformer, where the
accuracies of the 8-512 and 8-768 models increase from 74.79\% and 75.85\%
to 75.96\% and 76.95\%, respectively. The consistent gains across both backbones and multiple model scales indicate that SLI--ACF remains effective under different spiking attention formulations and capacity settings.

\begin{table*}[!t]
\centering
\caption{Comparison on CIFAR-10, CIFAR-100, and CIFAR10-DVS.
The Spikingformer and QKFormer baselines in the lower block are reproduced under a unified benchmark for direct comparison with their ACF-enhanced counterparts.}
\label{tab:cifar}
\setlength{\tabcolsep}{5pt}
\begin{tabular}{lccc ccc ccc}
\toprule
\multirow{2}{*}{Method}
& \multicolumn{3}{c}{\textbf{CIFAR10}}
& \multicolumn{3}{c}{\textbf{CIFAR100}}
& \multicolumn{3}{c}{\textbf{CIFAR10-DVS}} \\
\cmidrule(lr){2-4} \cmidrule(lr){5-7} \cmidrule(lr){8-10}
& Params (M) & $T$ & Acc. (\%)
& Params (M) & $T$ & Acc. (\%)
& Params (M) & $T$ & Acc. (\%) \\
\midrule


$\alpha$-SSA~\cite{xiao2025rethinking}
    & 9.32 & 4 & 95.97 & 9.32 & 4 & 80.22 & 2.57 & 16 & 81.64 \\
SacSSA~\cite{wangspiking}
    & 5.57 & 4 & 96.1 & 5.57 & 4 & 80.1 & 1.52 & 16 & 82.3 \\
$A^2OS^2$~\cite{guo2025spiking}
    & 10.23 & 4 & 96.42 & 10.23 & 4 & 79.90 & - & - & -\\
STAtten+~\cite{lee2025spiking}
    & - & 4 & 95.35 & - & 4 & 80.20 & - & 16 & 83.9   \\
SEMM~\cite{lee2025spiking}
    & - & 4 & 96.16 & - & 4 & 80.24 & - & 16 & 82.1   \\
TEFormer~\cite{shen2026teformer} 
    & 7.77 & 4 &  96.24  & 7.77 & 4 & 79.84  & 3.81 & 10 & 81.9 \\
SDT~\cite{sdtv1}
    & 9.32 & 4 & 95.60 & 9.32 & 4 & 78.40 & 2.57 & 16 & 80.0\\
Spikformer~\cite{zhou2022spikformer}
    & 9.32 & 4 & 95.51 & 9.32 & 4 & 78.21 & 2.57 & 16 & 80.9\\
CML~\cite{zhou2023enhancing}
    & 9.32 & 4 & 95.95 & 9.32 & 4 & 80.37 & 2.57 & 16 & 81.4 \\
\midrule
Spikingformer~\cite{zhou2023spikingformer}
    & 9.32 & 4 & 95.75 & 9.32 & 4 & 79.96 & 2.57 & 16 & 80.9   \\
QKFormer~\cite{zhou2024qkformer}
    & 6.74 & 4 & 96.25 & 6.74 & 4 & 81.15 & 1.50 & 16 & 83.0 \\
\midrule

Spikingformer + \textbf{SLI+ACF}
    & 9.93 & 4 & \textbf{96.26} & 9.93 & 4 & \textbf{80.84} & 2.70 & 16 & \textbf{82.3} \\
QKFormer + \textbf{SLI+ACF}
    & 7.06 & 4 & \textbf{96.55} & 7.06 & 4 & \textbf{81.75} & 1.59& 16 & \textbf{84.0} \\
    
\bottomrule
\end{tabular}

\end{table*}

\subsection{Results on CIFAR and Neuromorphic Datasets}

We further evaluate the proposed method on
frame-based and event-based benchmarks. Table~\ref{tab:cifar} reports the
results on CIFAR-10, CIFAR-100~\cite{krizhevsky2009learning}, and CIFAR10-DVS~\cite{li2017cifar10}. Under matched architectures
and temporal settings, incorporating SLI and ACF consistently improves both
Spikingformer and QKFormer.

On the frame-based datasets, Spikingformer achieves 96.26\% on CIFAR-10 and
80.84\% on CIFAR-100, compared with 95.75\% and 79.96\% for the reproduced
baseline. The corresponding QKFormer results increase from 96.25\% and
81.15\% to 96.55\% and 81.75\%, respectively. QKFormer equipped with SLI and
ACF achieves the highest accuracies listed for both datasets.

We further evaluate the method on CIFAR10-DVS to examine its applicability
to temporally evolving event streams. At $T=16$, Spikingformer achieves
82.3\%, compared with 80.9\% for the baseline, while QKFormer increases from
83.0\% to 84.0\%. The latter obtains the highest CIFAR10-DVS accuracy in
Table~\ref{tab:cifar} with 1.59M parameters. These results demonstrate
consistent effectiveness across frame-based and event-based inputs and
compatibility with different spiking attention backbones.

\subsection{Semantic Segmentation on ADE20K}

We evaluate SLI--ACF on ADE20K~\cite{zhou2017scene} to examine its applicability to dense visual prediction. Following SDT-V2~\cite{yao2024spike}, we adopt an FPN-based segmentation framework with a simulation length of $T=4$. Both Spikingformer and QKFormer employ eight-layer backbones with an embedding dimension of 512 and are trained without ImageNet pretraining.

Table~\ref{tab:ade20k} shows that SLI--ACF improves the mIoU of Spikingformer from $31.6\%$ to $34.5\%$ and that of QKFormer from $32.6\%$ to $37.5\%$, with only $0.6$M additional parameters for each backbone. In particular, QKFormer with SLI--ACF achieves the highest mIoU among the compared Transformer-based SNNs, reaching $37.5\%$ with 14.2M parameters. It also outperforms the 58.9M-parameter SDT-V2 at the same simulation length, despite using substantially fewer parameters and no ImageNet pretraining. The consistent gains on both backbones indicate that local spatial interaction provides an effective complement to spiking attention for dense prediction.

\begin{table}[!t]
\centering
\caption{Semantic segmentation performance on ADE20K.}
\label{tab:ade20k}
\small
\setlength{\tabcolsep}{3pt}
\renewcommand{\arraystretch}{1.05}

\begin{adjustbox}{max width=\columnwidth}
\begin{tabular}{@{}lcccc@{}}
\toprule
Method & Pretrain & Params (M) & $T$ & mIoU (\%) \\
\midrule

\multicolumn{5}{@{}l}{\textit{ANN}} \\
\cmidrule(lr){1-5}
ResNet-18~\cite{yu2022metaformer}
& $\checkmark$ & 15.5 & 1 & 32.9 \\
PVT-Tiny~\cite{wang2021pyramid}
& $\checkmark$ & 17.0 & 1 & 35.7 \\
PVT-Small~\cite{wang2021pyramid}
& $\checkmark$ & 28.2 & 1 & 39.8 \\
DeepLab-V3~\cite{zhang2022resnest}
& $\checkmark$ & 68.1 & 1 & 42.7 \\

\midrule
\multicolumn{5}{@{}l}{\textit{Transformer-based SNN}} \\
\cmidrule(lr){1-5}
\multirow{4}{*}{SDT-V2~\cite{yao2024spike}}
& $\checkmark$ & 16.5 & 1 & 32.3 \\
& $\checkmark$ & 16.5 & 4 & 33.6 \\
\cmidrule(lr){2-5}
& $\checkmark$ & 58.9 & 1 & 34.8 \\
& $\checkmark$ & 58.9 & 4 & 35.3 \\

\midrule
Spikingformer
& $\times$ & 13.8 & 4 & 31.6 \\
Spikingformer + \textbf{SLI+ACF}
& $\times$ & 14.4 & 4 & \textbf{34.5} \\

\midrule
QKFormer
& $\times$ & 13.6 & 4 & 32.6 \\
QKFormer + \textbf{SLI+ACF}
& $\times$ & 14.2 & 4 & \textbf{37.5} \\

\bottomrule
\end{tabular}
\end{adjustbox}
\end{table}
\begin{figure}[!ht]
    \centering
    \includegraphics[width=\linewidth]{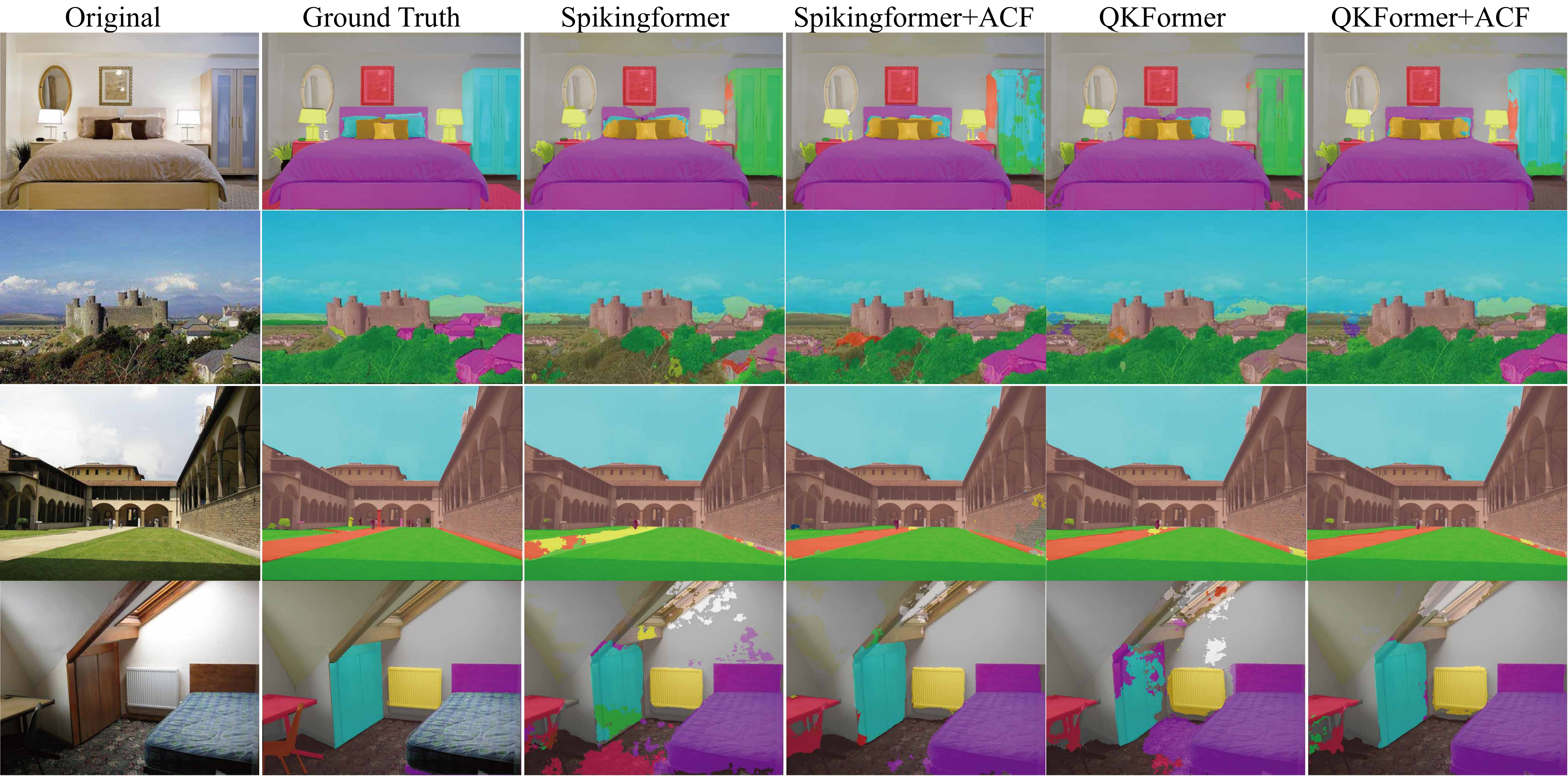}
\caption{
Qualitative ADE20K results of Spikingformer and QKFormer with and without SLI and ACF.
}
    \label{fig:seg_qk}
\end{figure}

The qualitative results in Fig.~\ref{fig:seg_qk} show that SLI--ACF produces more spatially coherent predictions for both backbones, with improved consistency over contiguous semantic regions and clearer delineation of object boundaries. These observations are consistent with the quantitative improvements reported in Table~\ref{tab:ade20k}.

\subsection{Ablation and Analysis}
\label{sec:ablation_analysis}

\paragraph{Component Complementarity and Initialization Stability.}

\begin{figure}[!t]
    \centering
    \includegraphics[width=\linewidth]{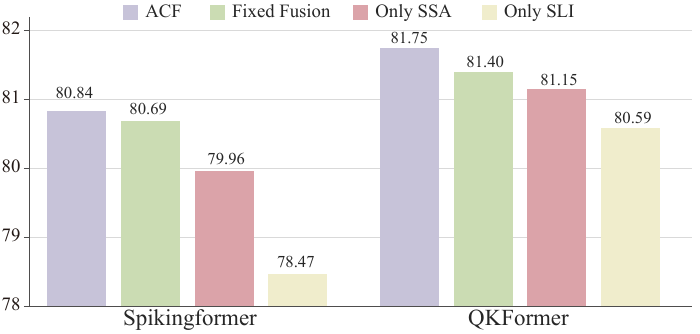}
    \caption{
    Component ablation on CIFAR-100 using
    Spikingformer-4-384 and QKFormer-4-384.
    }
    \label{fig:component_ablation}
\end{figure}

We compare four interaction configurations, including SSA only, SLI only,
fixed fusion with
$\gamma_{\mathrm{ssa}}=\gamma_{\mathrm{sli}}=0.5$, and learnable ACF.
As shown in Fig.~\ref{fig:component_ablation}, combining SSA and SLI
consistently yields higher accuracy than either pathway alone on both
Spikingformer and QKFormer. The improvement achieved by fixed fusion suggests that attention-based relations and local spatial interactions capture complementary information. Further gains from channel-wise learnable coefficients show that their relative contributions are better determined adaptively than fixed in advance.

We further evaluate the sensitivity of the fusion coefficients to their
initial values by varying a shared coefficient
$c\in\{0.1,0.25,0.5,0.75,1.0\}$. For fixed fusion, both coefficients remain at $c$, whereas for learnable fusion they are initialized to $c$ and optimized during training.
Table~\ref{tab:gamma_analysis} reports the mean and standard deviation over
the five initializations. Across both backbones, learnable fusion achieves higher mean accuracy and substantially lower variation, indicating reduced sensitivity to coefficient initialization.

\begin{table}[!htbp]
    \centering
\caption{
Mean and standard deviation of Top-1 accuracy across five fusion-coefficient
settings on CIFAR-100.
}
    \label{tab:gamma_analysis}

    \begin{tabular}{lcccc}
        \toprule
        \multirow{2}{*}{Backbone}
        & \multicolumn{2}{c}{Mean Top-1 (\%) $\uparrow$}
        & \multicolumn{2}{c}{Std. $\downarrow$} \\
        \cmidrule(lr){2-3}
        \cmidrule(lr){4-5}
        & Fixed
        & Learnable
        & Fixed
        & Learnable \\
        \midrule
        QKFormer
        & 81.64
        & \textbf{81.74}
        & 0.33
        & \textbf{0.08} \\
        Spikingformer
        & 80.62
        & \textbf{80.77}
        & 0.17
        & \textbf{0.06} \\
        \bottomrule
    \end{tabular}
\end{table}

\paragraph{Learned Fusion Patterns across Datasets and Backbones.}

\begin{figure}[t]
    \centering
    \includegraphics[width=\linewidth]{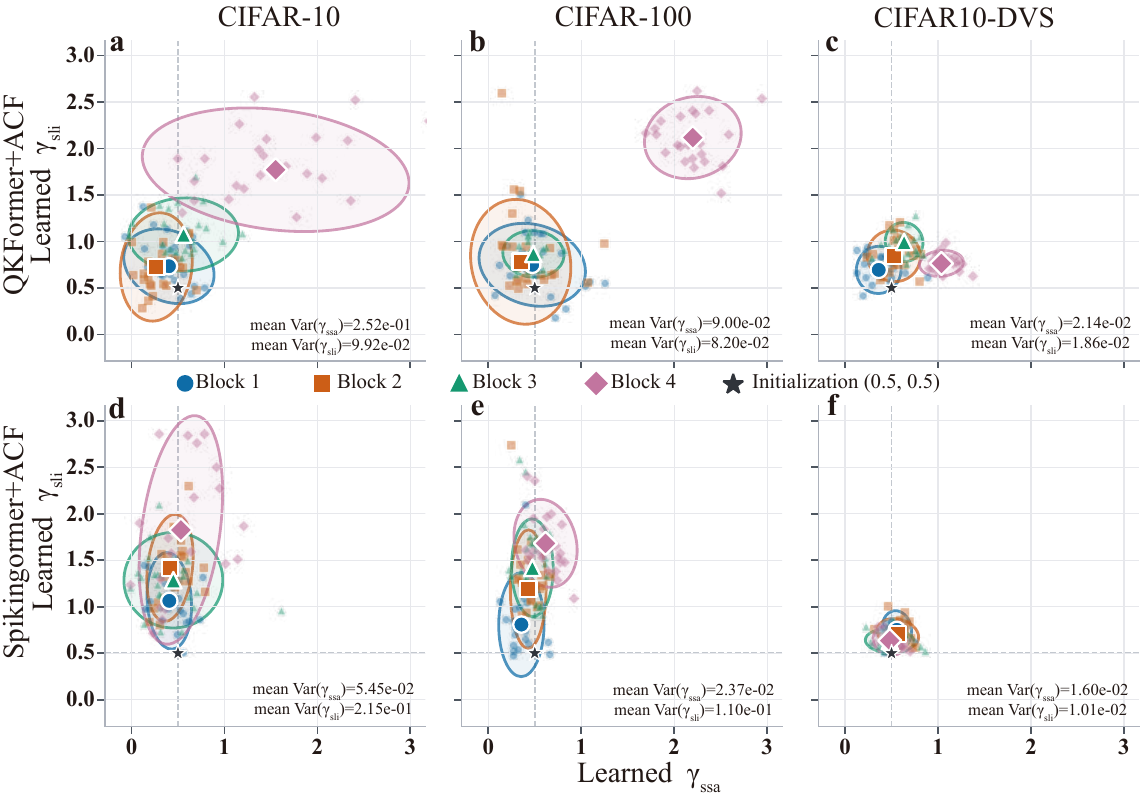}
    \caption{
Channel-wise distributions of the learned
$\gamma_{\mathrm{ssa}}$ and $\gamma_{\mathrm{sli}}$
across blocks, backbones, and datasets.
    }
    \label{fig:gamma_distribution}
\end{figure}

Figure~\ref{fig:gamma_distribution} shows that the channel-wise coefficients $\gamma_{\mathrm{ssa}}$ and $\gamma_{\mathrm{sli}}$, although uniformly initialized to $0.5$, develop distinct distributions after training. On CIFAR-10 and CIFAR-100, the final block of QKFormer differs markedly from the preceding blocks along both coefficient dimensions, whereas the block-wise variation in Spikingformer is more pronounced for $\gamma_{\mathrm{sli}}$. On CIFAR10-DVS, the learned distributions are more compact and exhibit greater overlap across blocks. These patterns suggest that ACF adjusts the relative contributions of SSA and SLI according to the backbone, network depth, and dataset rather than converging to a uniform fusion scheme.

\paragraph{Branch-Specific Spatial Responses.}

\begin{figure}[!htbp]
    \centering
    \includegraphics[width=\linewidth]{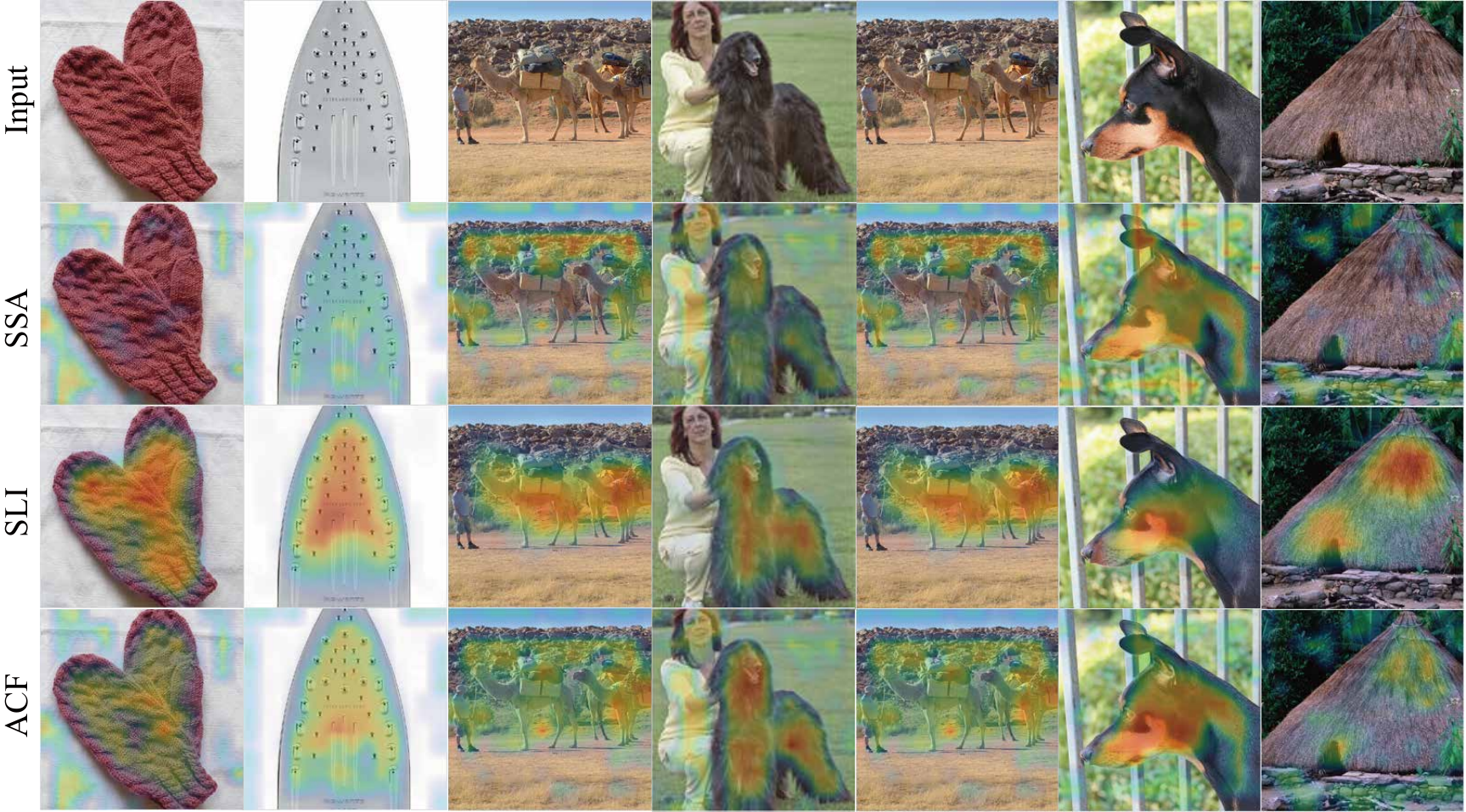}
    \caption{
    Grad-CAM++ visualizations of SSA, SLI, and their ACF-fused responses
    on ImageNet-1K.
    }
    \label{fig:gradcam_imagenet1}
\end{figure}

The Grad-CAM++ visualizations~\cite{chattopadhay2018grad} in
Fig.~\ref{fig:gradcam_imagenet1} show distinct spatial response patterns for
SSA and SLI. SSA produces relatively broad responses that cover multiple
object parts and surrounding context, whereas SLI yields more localized
responses over spatially contiguous object regions. Their fusion through ACF
preserves concentrated responses on the target object while retaining
contextual information captured by SSA. These observations are consistent with the ablation results and indicate complementary spatial responses between the two branches.

\section{Conclusion}

We presented Spiking Local Interaction (SLI) and Adaptive Complementary Fusion (ACF) to enhance token interaction in Spiking Transformers. SLI introduces an attention-independent pathway for local information exchange, complementing the input-dependent relations modeled by spiking self-attention. ACF further learns layer-specific, channel-wise coefficients to balance the two interaction pathways according to the evolving representations across network depth. The resulting design can be integrated into existing Spiking Transformer architectures with modest parameter overhead. Experiments across image classification, event-based recognition, and semantic segmentation show consistent improvements on different backbones and model scales. The proposed method achieves state-of-the-art performance on ImageNet-1K, CIFAR-10, CIFAR-100, and CIFAR10-DVS, and obtains the highest mIoU among the compared Transformer-based SNNs on ADE20K. Ablation studies and qualitative analyses further show that SSA and SLI capture complementary interaction patterns, while adaptive fusion provides a more effective combination than fixed weighting. These results establish local interaction as a useful complement to spiking attention and provide a simple approach for improving token representations in Spiking Transformers.

\bibliography{aaai2027}


\clearpage
\section{Architecture Details}
\label{app:architecture}

\subsection{Overall Architecture}
\label{app:overall_architecture}

We evaluate SLI and ACF on two representative Spiking Transformer backbones,
Spikingformer and QKFormer. For image classification, both backbones comprise
a convolutional spiking tokenizer, stacked Transformer encoder blocks, global
average pooling, and a linear classification head. For semantic segmentation,
they are adapted to a hierarchical feature-pyramid architecture.

Static images are repeated over $T$ simulation steps, whereas neuromorphic
event streams are integrated into $T$ frames. Unless otherwise specified,
LIF neurons use $\tau=2$ and $V_{\mathrm{th}}=1$, while those applied to
attention responses use $V_{\mathrm{th}}=0.5$. For SSA, attention outputs are scaled
by $s=0.125$, whereas QKTA does not use a scaling factor. Batch normalization follows convolutional and linear
projection layers within the backbone.


\subsection{Classification Backbones}
\label{app:classification_backbones}

\subsubsection{Spikingformer}
\label{app:spikingformer}

Spikingformer adopts a columnar architecture with a constant embedding
dimension $D$ across all $L$ Transformer encoder blocks. Its spiking tokenizer
contains five $3\times3$ convolutional layers with channel dimensions
$D/8 \rightarrow D/4 \rightarrow D/2 \rightarrow D \rightarrow D$. For
CIFAR-scale inputs, the tokenizer produces an $8\times8$ token map, whereas
ImageNet-1K inputs are reduced to a $14\times14$ token map. In the original Spikingformer architecture, each block consists of an SSA
sub-block followed by a spiking MLP.

\subsubsection{QKFormer}
\label{app:qkformer}

QKFormer adopts a hierarchical architecture in which the spatial resolution
decreases and the embedding dimension increases across stages. For
ImageNet-1K, it forms a three-stage hierarchy with spatial resolutions
$56^2 \rightarrow 28^2 \rightarrow 14^2$, channel dimensions
$D/4 \rightarrow D/2 \rightarrow D$, and block allocations of $1$, $2$, and
$L-3$. For CIFAR-scale inputs, the corresponding hierarchy is
$32^2 \rightarrow 16^2 \rightarrow 8^2$, with block allocations of $1$, $1$,
and $L-2$. Following the original QKFormer design, the high-resolution stages
use Token Q--K Attention, while the low-resolution stage uses SSA. Each
patch-embedding module additionally contains a $1\times1$ stride-$2$ convolutional
shortcut. The ImageNet-scale configurations are summarized in
Table~\ref{tab:app_imagenet_architecture}.


\subsection{Integration of SLI and ACF}
\label{app:sli_acf_integration}

SLI is introduced in parallel with the attention sub-block of each Transformer
block. Given the block input, the attention pathway models input-dependent
token relations through SSA or TSSA, while SLI exchanges information among
neighboring tokens through a lightweight depthwise--pointwise transformation.
ACF then combines the two pathways using the layer-specific, channel-wise
scaling vectors $\gamma_{\mathrm{ssa},l}$ and
$\gamma_{\mathrm{sli},l}$. The fused representation is subsequently processed
by the standard spiking MLP. The same integration strategy is applied to both Spikingformer and QKFormer
without modifying their original tokenizers, attention formulations, MLP
sub-blocks, or prediction heads. All entries of
$\gamma_{\mathrm{ssa},l}$ and $\gamma_{\mathrm{sli},l}$ are initialized to
$0.5$.

\subsection{Classification Model Configurations}
\label{app:classification_configurations}

We denote a model containing $L$ Transformer blocks and a maximum
embedding dimension of $D$ as $L$-$D$. For CIFAR-10 and CIFAR-100, we use
Spikingformer-4-384 with $12$ attention heads and QKFormer-4-384 with $8$
attention heads. QKFormer-4-384 adopts stage widths of $96/192/384$ and block
allocations of $1/1/2$. Static images are simulated for $T=4$ time steps. The ImageNet-scale configurations are summarized in
Table~\ref{tab:app_imagenet_architecture}. Both Spikingformer and QKFormer use 
$8$ attention heads on ImageNet-1K. For CIFAR10-DVS, the corresponding backbone variants receive
$128\times128$ two-channel event frames and operate for $T=16$ time steps.

\begin{table*}[t]
  \centering
  \small
  \renewcommand{\arraystretch}{1.25}
  \setlength{\tabcolsep}{6pt}
  \begin{tabular}{|c|c|c|c|c|c|c|}
    \hline
    Stage
      & Spatial resolution
      & \multicolumn{2}{c|}{Layer specification}
      & 8-512
      & 10-512
      & 8-768 \\
    \hline

    \multicolumn{7}{|c|}{\emph{Spikingformer (columnar)}} \\
    \hline

    \multirow{5}{*}{\shortstack{Spiking\\Tokenizer}}
      & $H\times W$
      & Downsampling 1
      & Conv $3\times3$
      & $64$
      & ---
      & $96$ \\
    \cline{2-7}

      & $\frac{H}{2}\times\frac{W}{2}$
      & Downsampling 2
      & MP + Conv $3\times3$
      & $128$
      & ---
      & $192$ \\
    \cline{2-7}

      & $\frac{H}{4}\times\frac{W}{4}$
      & Downsampling 3
      & MP + Conv $3\times3$
      & $256$
      & ---
      & $384$ \\
    \cline{2-7}

      & $\frac{H}{8}\times\frac{W}{8}$
      & Downsampling 4
      & MP + Conv $3\times3$
      & $512$
      & ---
      & $768$ \\
    \cline{2-7}

      & $\frac{H}{16}\times\frac{W}{16}$
      & Downsampling 5
      & MP + Conv $3\times3$
      & $512$
      & ---
      & $768$ \\
    \hline

    \multirow{2}{*}{\shortstack{SSA\\Blocks}}
      & \multirow{2}{*}
        {$\frac{H}{16}\times\frac{W}{16}$}
      & Block type
      & SSA block
      & \multicolumn{3}{c|}{} \\
    \cline{3-7}

      & & Number of blocks
      &
      & $8$
      & ---
      & $8$ \\
    \hline

    \multicolumn{7}{|c|}{\emph{QKFormer (hierarchical)}} \\
    \hline

    \multirow{2}{*}{1}
      & \multirow{2}{*}
        {$\frac{H}{4}\times\frac{W}{4}$}
      & Downsampling
      & Conv $3\times3$, MP $\times2$
      & $128$
      & $128$
      & $192$ \\
    \cline{3-7}

      & & Blocks
      & TSSA block
      & $\times1$
      & $\times1$
      & $\times1$ \\
    \hline

    \multirow{2}{*}{2}
      & \multirow{2}{*}
        {$\frac{H}{8}\times\frac{W}{8}$}
      & Downsampling
      & Conv $3\times3$, MP
      & $256$
      & $256$
      & $384$ \\
    \cline{3-7}

      & & Blocks
      & TSSA block
      & $\times2$
      & $\times2$
      & $\times2$ \\
    \hline

    \multirow{2}{*}{3}
      & \multirow{2}{*}
        {$\frac{H}{16}\times\frac{W}{16}$}
      & Downsampling
      & Conv $3\times3$, MP
      & $512$
      & $512$
      & $768$ \\
    \cline{3-7}

      & & Blocks
      & SSA block
      & $\times5$
      & $\times7$
      & $\times5$ \\
    \hline

    \multicolumn{7}{|c|}
      {\emph{Transformer block specifications}} \\
    \hline

    \multirow{4}{*}{\shortstack{TSSA\\Block}}
      & \multicolumn{2}{c|}{Attention pathway}
      & \multicolumn{4}{l|}
        {Token Q--K Attention with $1\times1$ Q/K projections} \\
    \cline{2-7}

      & \multicolumn{2}{c|}{SLI pathway}
      & \multicolumn{4}{l|}
        {DWConv $3\times3$ + PWConv $1\times1$ + BN + SN} \\
    \cline{2-7}

      & \multicolumn{2}{c|}{ACF}
      & \multicolumn{4}{l|}
        {Channel-wise fusion with learnable
        $\gamma_{\mathrm{ssa},l}$ and
        $\gamma_{\mathrm{sli},l}$} \\
    \cline{2-7}

      & \multicolumn{2}{c|}{Channel MLP}
      & \multicolumn{4}{l|}
        {Conv $1\times1$ $\times2$; expansion ratio $4$} \\
    \hline

    \multirow{4}{*}{\shortstack{SSA\\Block}}
      & \multicolumn{2}{c|}{Attention pathway}
      & \multicolumn{4}{l|}
        {Spiking self-attention with $1\times1$ Q/K/V projections} \\
    \cline{2-7}

      & \multicolumn{2}{c|}{SLI pathway}
      & \multicolumn{4}{l|}
        {DWConv $3\times3$ + PWConv $1\times1$ + BN + SN} \\
    \cline{2-7}

      & \multicolumn{2}{c|}{ACF}
      & \multicolumn{4}{l|}
        {Channel-wise fusion with learnable
        $\gamma_{\mathrm{ssa},l}$ and
        $\gamma_{\mathrm{sli},l}$} \\
    \cline{2-7}

      & \multicolumn{2}{c|}{Channel MLP}
      & \multicolumn{4}{l|}
        {Conv $1\times1$ $\times2$; expansion ratio $4$} \\
    \hline
  \end{tabular}

  \caption{
  Architecture specifications of the ImageNet-scale Spikingformer and
  QKFormer backbones with SLI and ACF. In each Transformer block, SLI
  operates in parallel with SSA or TSSA, and the two pathways are fused
  by ACF before the spiking MLP. MP denotes $3\times3$ max-pooling with
  stride $2$, and each QKFormer downsampling module includes a
  $1\times1$ stride-$2$ convolutional shortcut.
  }
  \label{tab:app_imagenet_architecture}
\end{table*}

\subsection{Segmentation Backbones for ADE20K}
\label{app:segmentation_backbones}

For semantic segmentation on ADE20K, we adopt the hierarchical backbone
design of SDT-V2~\cite{yao2024spike} to produce multi-scale feature
representations. The input is processed by a $7\times7$ stride-$2$ spiking convolution,
followed by three stride-$2$ downsampling stages with channel dimensions of
$64$, $128$, and $256$. A subsequent stride-$1$ convolution expands the
channel dimension to $360$.

The high-resolution stages use convolution-based SNN blocks with separable
spatial convolution, channel mixing, and membrane shortcuts. The
low-resolution stages contain six Transformer blocks at width $256$ and two
at width $360$. Spikingformer uses SSA throughout, whereas QKFormer employs
TSSA in the first three blocks and SSA in the remaining five. 

SLI and ACF are applied to all Transformer blocks, while the convolution-based
stages retain their original structure. The backbone provides a four-level
feature hierarchy $\{P_1,P_2,P_3,P_4\}$ with channel dimensions
$\{32,64,128,360\}$ and output strides $\{2,4,8,16\}$, respectively.

\begin{table*}[t]
  \centering
  \small
  \renewcommand{\arraystretch}{1.25}
  \setlength{\tabcolsep}{6pt}
  \begin{tabular}{|c|c|c|c|c|c|}
    \hline
    Stage
      & Spatial resolution
      & \multicolumn{2}{c|}{Layer specification}
      & Spikingformer
      & QKFormer \\
    \hline

    \multirow{2}{*}{1}
      & \multirow{2}{*}{$\frac{H}{2}\times\frac{W}{2}$}
      & Downsampling
      & Conv $7\times7$, stride $2$, dim $32$
      & \multicolumn{2}{c|}{} \\
    \cline{3-6}

      & & Blocks
      & Conv-based SNN block
      & $\times1\rightarrow P_1$
      & $\times1\rightarrow P_1$ \\
    \hline

    \multirow{2}{*}{2}
      & \multirow{2}{*}{$\frac{H}{4}\times\frac{W}{4}$}
      & Downsampling
      & Conv $3\times3$, stride $2$, dim $64$
      & \multicolumn{2}{c|}{} \\
    \cline{3-6}

      & & Blocks
      & Conv-based SNN block
      & $\times1\rightarrow P_2$
      & $\times1\rightarrow P_2$ \\
    \hline

    \multirow{2}{*}{3}
      & \multirow{2}{*}{$\frac{H}{8}\times\frac{W}{8}$}
      & Downsampling
      & Conv $3\times3$, stride $2$, dim $128$
      & \multicolumn{2}{c|}{} \\
    \cline{3-6}

      & & Blocks
      & Conv-based SNN block
      & $\times2\rightarrow P_3$
      & $\times2\rightarrow P_3$ \\
    \hline

    \multirow{2}{*}{4}
      & \multirow{2}{*}{$\frac{H}{16}\times\frac{W}{16}$}
      & Downsampling
      & Conv $3\times3$, stride $2$, dim $256$
      & \multicolumn{2}{c|}{} \\
    \cline{3-6}

      & & Blocks
      & Transformer SNN block
      & SSA $\times6$
      & TSSA $\times3$ + SSA $\times3$ \\
    \hline

    \multirow{2}{*}{5}
      & \multirow{2}{*}{$\frac{H}{16}\times\frac{W}{16}$}
      & Channel expansion
      & Conv $3\times3$, stride $1$, dim $360$
      & \multicolumn{2}{c|}{} \\
    \cline{3-6}

      & & Blocks
      & Transformer SNN block
      & SSA $\times2\rightarrow P_4$
      & SSA $\times2\rightarrow P_4$ \\
    \hline

    \multicolumn{6}{|c|}{\emph{Block specifications}} \\
    \hline

    \multirow{2}{*}{\shortstack{Conv-based\\SNN Block}}
      & \multicolumn{2}{c|}{SepConv}
      & \multicolumn{3}{l|}
      {PW $1\times1$ (ratio $2$)
      $\rightarrow$ DW $7\times7$
      $\rightarrow$ PW $1\times1$;
      membrane shortcut} \\
    \cline{2-6}
    
      & \multicolumn{2}{c|}{Channel convolution}
      & \multicolumn{3}{l|}
      {Conv $3\times3$ $\times2$;
      expansion ratio $4$;
      membrane shortcut} \\
    \hline

    \multirow{4}{*}{\shortstack{Transformer\\SNN Block}}
      & \multicolumn{2}{c|}{Attention pathway}
      & \multicolumn{3}{l|}
      {SSA or TSSA with $8$ attention heads} \\
    \cline{2-6}

      & \multicolumn{2}{c|}{SLI pathway}
      & \multicolumn{3}{l|}
      {DWConv $3\times3$ + PWConv $1\times1$ + BN + SN} \\
    \cline{2-6}

      & \multicolumn{2}{c|}{ACF}
      & \multicolumn{3}{l|}
      {Channel-wise fusion with learnable
      $\gamma_{\mathrm{ssa},l}$ and
      $\gamma_{\mathrm{sli},l}$} \\
    \cline{2-6}

      & \multicolumn{2}{c|}{Channel MLP}
      & \multicolumn{3}{l|}
      {Conv $1\times1$ $\times2$;
      expansion ratio $4$} \\
    \hline
  \end{tabular}

\caption{
Architecture specifications of the ADE20K segmentation backbones with
SLI and ACF. In each Transformer SNN block, SLI operates in parallel
with SSA or TSSA, and their outputs are fused by ACF before the spiking
MLP. $P_1$--$P_4$ denote the multi-scale features passed to the FPN neck.
The downsampling modules follow SDT-V2 and contain convolutional
shortcuts. PW and DW denote pointwise and depthwise convolutions,
respectively.
}
  \label{tab:app_ade20k_architecture}
\end{table*}


\section{Experimental Settings}
\label{app:experimental_settings}

For each backbone, the baseline and its SLI--ACF variant follow identical
data-processing, optimization, and evaluation protocols, enabling a
controlled assessment of the proposed components.

\subsection{Datasets and Evaluation Metrics}
\label{app:datasets}

\paragraph{CIFAR-10 and CIFAR-100.}
CIFAR-10 and CIFAR-100~\cite{krizhevsky2009learning} are static-image
classification benchmarks comprising $60{,}000$ RGB images at a resolution
of $32\times32$, including $50{,}000$ training images and $10{,}000$ test
images. CIFAR-10 contains $10$ categories, whereas CIFAR-100 contains $100$
fine-grained categories. All models are trained from scratch, and Top-1 accuracy is reported on the
official test set of each dataset.

\paragraph{CIFAR10-DVS.}
CIFAR10-DVS~\cite{li2017cifar10} is a neuromorphic classification benchmark
constructed by recording moving CIFAR-10 images with a dynamic vision sensor.
It contains $10{,}000$ event streams from $10$ categories. Following the commonly adopted protocol, the samples are split into training
and test sets at a $9{:}1$ ratio. Each event stream is integrated into $T=16$ frames with two
polarity channels at a spatial resolution of $128\times128$. Performance is
evaluated using Top-1 accuracy on the test set.

\paragraph{ImageNet-1K.}
ImageNet-1K~\cite{deng2009imagenet} contains approximately $1.28$ million
training images and $50{,}000$ validation images spanning $1{,}000$ object
categories. All models are trained with an input resolution of
$224\times224$, and Top-1 accuracy is reported on the validation set.

\paragraph{ADE20K.}
ADE20K~\cite{zhou2017scene} is a semantic segmentation benchmark containing
$20{,}210$ training images and $2{,}000$ validation images with
pixel-level annotations for $150$ categories. All segmentation
models are trained from scratch without ImageNet pretraining, and mean
intersection over union (mIoU) is reported on the validation set.

\subsection{Classification Training Details}
\label{app:classification_training} 

All classification models are optimized using AdamW with cosine learning-rate
decay and linear warmup. We employ label smoothing with a coefficient of
$0.1$, automatic mixed-precision training, and a sigmoid surrogate-gradient
function with $\alpha=4$. The complete training configurations are reported
in Table~\ref{tab:app_hyper}.

\begin{table}[t]
  \centering
  \small
  \renewcommand{\arraystretch}{1.15}
  \setlength{\tabcolsep}{3pt}
  \begin{tabular}{lccc}
    \toprule
    Configuration & CIFAR-10/100 & CIFAR10-DVS & ImageNet-1K \\
    \midrule
    Input size        & $32^2$  & $128^2$ & $224^2$ \\
    Time steps $T$    & $4$     & $16$    & $4$ \\
    Epochs            & $400$   & $96$    & $300$ / $200$ \\
    Warmup epochs     & $20$    & $10$    & $20$ / $5$ \\
    Batch size        & $64$    & $16$    & $24$ per GPU \\
    Learning rate     & $10^{-3}$ & $10^{-3}$ & $5 \times 10^{-4}$ \\
    Minimum LR        & $10^{-5}$ & $10^{-5}$ & $10^{-5}$ \\
    Weight decay      & $0.06$  & $0.06$  & $0.05$ \\
    RandAugment       & m9-n1   & ---     & m9 \\
    Mixup             & $0.5$   & $0.5$   & $0.8$ \\
    CutMix            & ---     & ---     & $1.0$ \\
    Random erasing    & $0.25$  & ---     & $0.25$ \\
    \bottomrule
  \end{tabular}
\caption{
Training configurations for the classification experiments. Values separated
by ``/'' correspond to Spikingformer and QKFormer, respectively. The reported
epoch counts exclude the subsequent $10$-epoch cooldown phase.
}
  \label{tab:app_hyper}
\end{table}

\paragraph{CIFAR-10 and CIFAR-100.}
Training augmentation includes random cropping, horizontal flipping,
RandAugment with magnitude $9$ and one operation per image, random erasing
with a probability of $0.25$, and Mixup with $\alpha=0.5$. Mixup is disabled
after epoch $200$. For each backbone, the reproduced baseline and its
SLI--ACF variant follow identical training settings. All experiments are
conducted on a single GPU.

\paragraph{CIFAR10-DVS.}
As conventional photometric transformations are not directly applicable to
event frames, we employ random horizontal flipping and the neuromorphic
AutoAugment policy adopted in prior Spiking Transformer studies. Mixup with
$\alpha=0.5$ is applied with a probability of $0.5$.

\paragraph{ImageNet-1K.}
Spikingformer variants are trained for $300$ epochs with $20$ warmup epochs,
whereas QKFormer variants follow the original $200$-epoch schedule with $5$
warmup epochs. The augmentation strategy includes RandAugment, Mixup, CutMix,
and random erasing. Training is performed on eight NVIDIA A100 GPUs with a
per-GPU batch size of $24$, yielding a total batch size of $192$. The configurations in Table~\ref{tab:app_hyper} apply to the principal
backbone experiments. Models denoted by $*$ in the main paper are reproduced
using their respective original training protocols.

\subsection{Semantic Segmentation Training Details}
\label{app:segmentation_training}

The ADE20K experiments are implemented within the
\texttt{EncoderDecoder} framework of MMSegmentation. The backbone outputs four feature
maps with channel dimensions of $32$, $64$, $128$, and $360$. These features
are averaged across $T=4$ simulation steps before being passed to the spike-driven FPN
neck and Semantic-FPN head adopted from SDT-V2~\cite{yao2024spike}. Each
intermediate convolution in the neck and segmentation head is preceded by
spike quantization of the temporally averaged features. The segmentation head predicts
$150$ semantic categories and is optimized using pixel-wise cross-entropy
loss.

All models are trained from scratch for $320$k iterations using
$512\times512$ random crops and a total batch size of $8$. AdamW is used with
an initial learning rate of $1\times10^{-3}$ and a weight decay of $0.005$.
The learning rate is linearly warmed up during the first $6$k iterations and
then decayed according to a polynomial schedule. A $2\times$ learning-rate
multiplier is applied to the FPN neck and segmentation head.

We additionally use synchronized batch normalization, gradient clipping, and
automatic mixed-precision training. We follow the standard ADE20K
augmentation protocol with random resizing, cropping, and horizontal
flipping. Evaluation is conducted on the validation set using single-scale whole-image
inference, with mIoU as the evaluation metric.

\end{document}